\documentclass[letterpaper, 10 pt, conference]{ieeeconf}  
\usepackage{graphicx}
\usepackage{caption}
\usepackage{amsfonts}
\usepackage{amsmath}
\usepackage{array}
\usepackage{makecell}
\usepackage{float}
\newcommand{\B}{\mathcal{B}}

\newcommand{\numDims}{d_{\mathcal{A}}}

\newcommand{\method}{\texttt{MoRE}}

\IEEEoverridecommandlockouts                              

\title{\LARGE \bf
\method: Unlocking Scalability in Reinforcement Learning for Quadruped Vision-Language-Action Models
}

\author{Han Zhao\textsuperscript{\ddag,1,2}, 
Wenxuan Song\textsuperscript{\ddag,3,4}, 
Donglin Wang\textsuperscript{\ddag,2*},
Xinyang Tong\textsuperscript{2},\\
Pengxiang Ding\textsuperscript{1,2}, 
Xuelian Cheng\textsuperscript{3}, 
Zongyuan Ge\textsuperscript{3} \\
\normalsize \textsuperscript{1}Zhejiang University, China
\normalsize \textsuperscript{2}MiLAB, Westlake University, China \\
\normalsize \textsuperscript{3}AIM Lab, Faculty of IT, Monash University, Australia
\normalsize \textsuperscript{4}HKUST(GZ), China \\}

\begin{document}

\maketitle
\thispagestyle{empty}
\pagestyle{empty}

\begin{abstract}
Developing versatile quadruped robots that can smoothly perform various actions and tasks in real-world environments remains a significant challenge. This paper introduces a novel vision-language-action (VLA) model, mixture of robotic experts (\method), for quadruped robots that aim to introduce reinforcement learning (RL) for fine-tuning large-scale VLA models with a large amount of mixed-quality data.
\method~integrates multiple low-rank adaptation modules as distinct experts within a dense multi-modal large language model (MLLM), forming a sparse-activated mixture-of-experts model. This design enables the model to effectively adapt to a wide array of downstream tasks. Moreover, we employ a reinforcement learning-based training objective to train our model as a Q-function after deeply exploring the structural properties of our tasks. Effective learning from automatically collected mixed-quality data enhances data efficiency and model performance. Extensive experiments demonstrate that \method~outperforms all baselines across six different skills and exhibits superior generalization capabilities in out-of-distribution scenarios. We further validate our method in real-world scenarios, confirming the practicality of our approach and laying a solid foundation for future research on multi-task learning in quadruped robots.

\end{abstract}
\renewcommand{\thefootnote}{}
\footnotetext{\ddag~Equal contribution}
\footnotetext{* Corresponding author. E-mail: wangdonglin@westlake.edu.cn}
\section{INTRODUCTION}

In the field of robotics research, enhancing the versatility and generalization of robots has remained a popular topic. These capabilities can effectively enhance the performance in following various language instructions to execute multiple tasks in complex and ever-changing open-world scenarios. 
With the rise of large language models~\cite{gpt4, llama, llama2} and visual perception research~\cite{clip, sam, xu2022multi, xu2024sdge}, there are numerous studies that leverage multi-modal large language models (MLLMs) as a controller for robots to handle different tasks, complex language instructions and visual scenes~\cite{palme,saycan}. MLLMs~\cite{llava, blip, fuyu-8b, cobra} learn extensive prior knowledge about images and natural language through large-scale pre-training on Internet-scale datasets, which have achieved significant development and application in numerous fields~\cite{lisa, xvila, pite}. 

Therefore, to combine the powerful vision-language understanding capabilities of MLLMs with the data-driven paradigm of robot learning~\cite{robovqa, openx}, many studies have been conducted to obtain vision-language-action (VLA) models by training these foundational models on data from the execution of various downstream tasks~\cite{brohan2023rt2,li2024roboflamingo,kim2024openvla, pdvla}, enabling these large models to directly receive visual signals and natural language commands and output executable action for robots in an end-to-end manner. However, most end-to-end large-scale VLA approaches primarily follow the following training and inference paradigms: 

\begin{figure}[t!]
    \centering
    \captionsetup{type=figure}
    \includegraphics[width=0.99\linewidth]{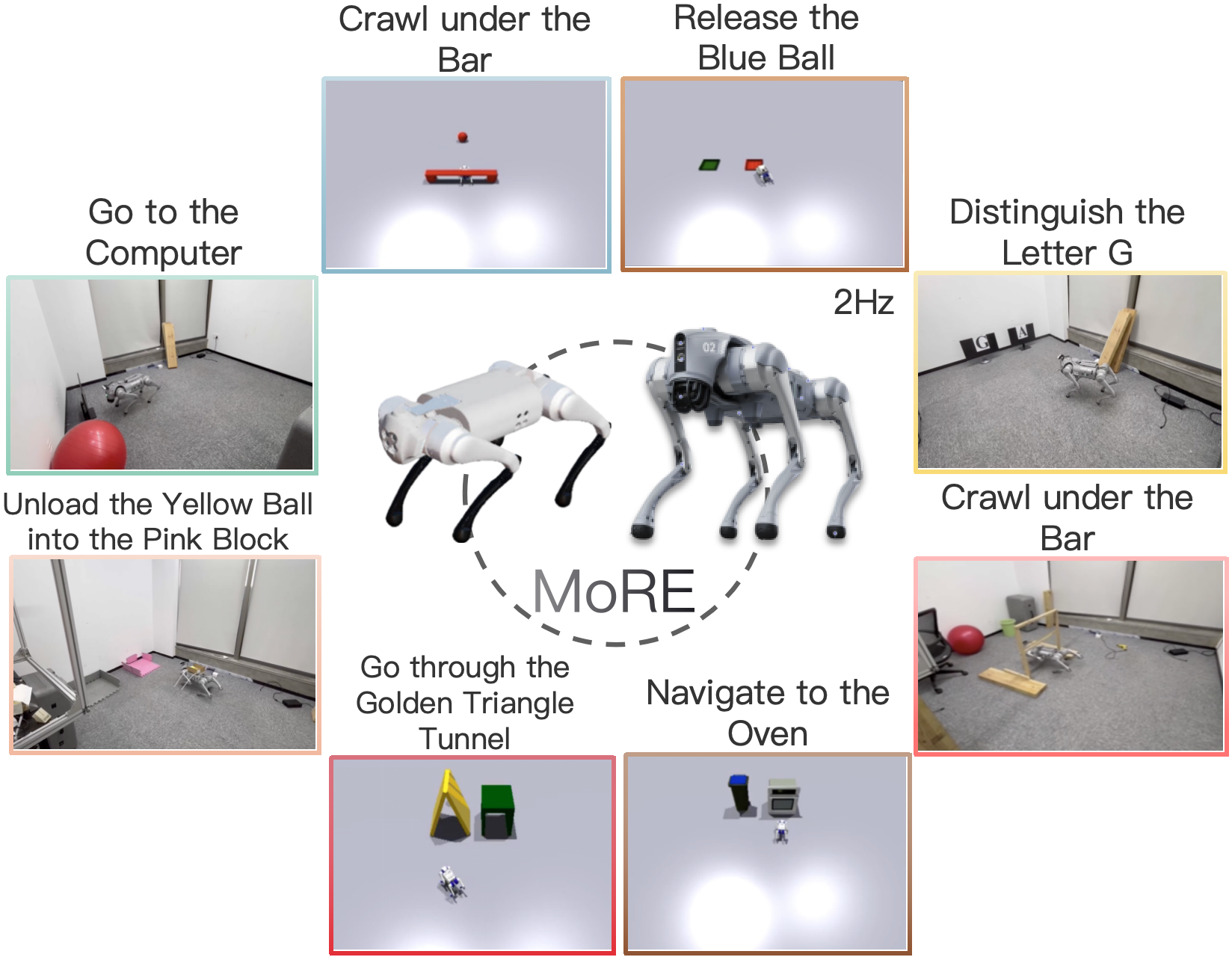}
    \vspace{-0.05in}
    \captionof{figure}{\textbf{Visualization of \method~on multi tasks.} \method~has been verified to be robust across various tasks, commands, and scenarios, in both simulation environments and real-world deployments.}
    \label{fig:visual}
    \vspace{-0.2in}
\end{figure}

1) These approaches typically involve directly fine-tuning on existing MLLM architectures, without delving deeper into whether such architectures are suitable for the diverse range of downstream robotic tasks. 

\begin{figure*}[t!]
    \centering
    \includegraphics[width=0.9\textwidth]{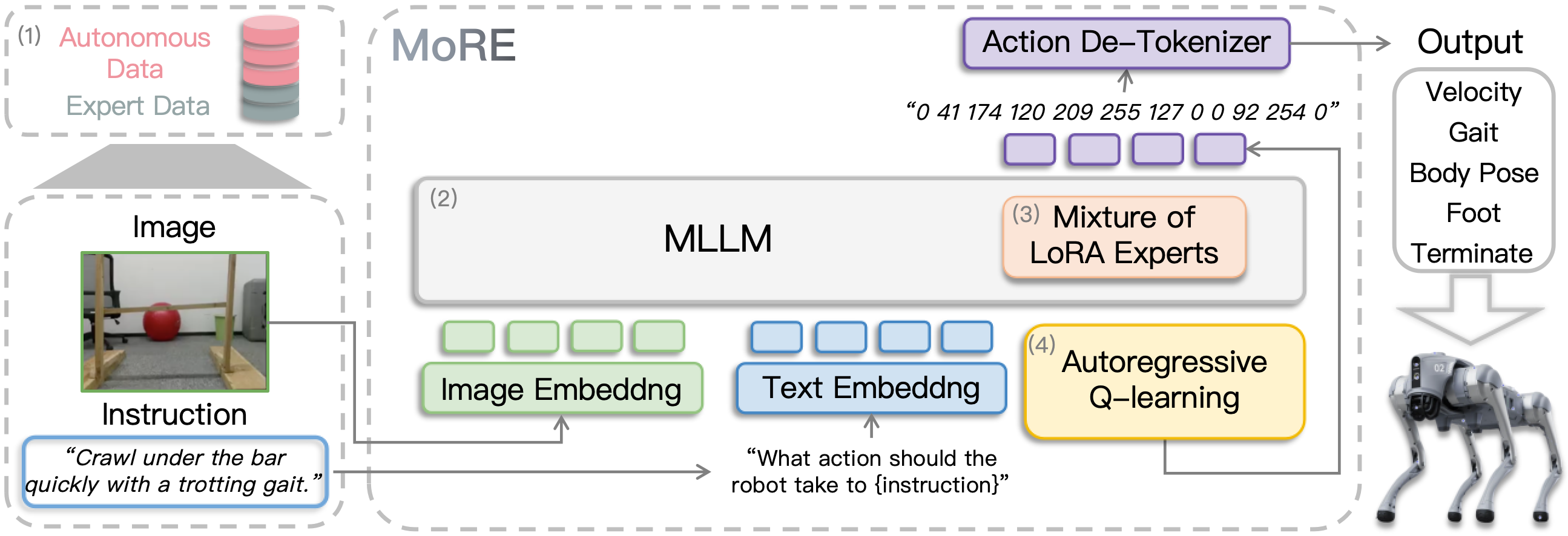}
    \caption{\textbf{Overview of \method} as applied to our multi-task quadruped vision-language-action task. 
    The overview consists of four key components: (1) broad \textbf{sub-optimal data} combined with narrow expert data, (2) the \textbf{MLLM backbone} to generate action tokens from image and text embedding, (3) the \textbf{Mixture of LoRA Experts} finetuned to adapt to different tasks, and (4) the \textbf{RL objectives} used for training.}
    \label{fig:overview}
\end{figure*}

2) Models based on the imitation learning (IL) paradigm are typically fine-tuned exclusively on expert data using a supervised training objective. However, collecting such expert data often requires substantial human supervision or complex automated data collection systems with data filtering~\cite{autort}. Consequently, these methods are unable to leverage more easily gathered sub-optimal data, which includes many trajectories that exhibit poor execution performance or even fail to complete tasks.

To address the aforementioned issues, we introduce Mixture of Robotic Experts (\method), a novel quadruped VLA model. \method~simultaneously receives RGB images and language instructions as inputs, and directly outputs control commands for quadruped robots. 
By incorporating multiple Low Rank Adaptation (LoRA) modules as different experts into a dense VLA model and fine-tuning it into a sparse-activated Mixture of LoRA Experts model, \method~achieves a balance for a wide variety of downstream quadruped robotic tasks. 
Additionally, it is widely understood that offline reinforcement learning (RL) is able to extract good policies from highly sub-optimal data, a scenario where imitation learning finds sub-optimal solutions that do not improve over the demonstrator that generated the dataset.
Thus, \method~employs a RL-based training objective to train our model as a Q-function after deeply exploring the structural properties of our task, effectively utilizing a large amount of automatically-collected mixed-quality data.

As a product of increased data diversity, new model components, and tailored training process, \method~outperforms all baselines by a clear margin across 6 skills.
Its high success rate in out-of-distribution experiments demonstrates superior generalization capabilities. 
Furthermore, we deployed our method in real-world scenarios, achieving promising trajectories. 
This confirms the practicality of our approach and lays a solid foundation for future research on multi-task learning in quadruped robots.

The main contributions in this paper include:
\begin{itemize}
    \item We propose \method, a novel VLA model for quadruped robots. To the best of our knowledge, this is the first work to explore the application of the MoE architecture in large-scale end-to-end VLA models. It shows high success rate and generalization in multi-task settings.
    \item By introducing a training objective based on RL, we have included sub-optimal trajectories collected automatically, thus effectively enhancing both data efficiency and model performance for large-scale VLA models.
    \item We conducted extensive experiments in both simulation and the real world to study the performance of \method~in various settings.
    
\end{itemize}

\section{RELATED WORKS}

\subsection{Vision-Language-Action Models}
Vision-language-action (VLA) models typically accept visual feedback from robots and operation instructions in natural language form as input, and directly outputs operation instructions that the robots can execute. Several large-scale VLA models~\cite{brohan2023rt2, kim2024openvla, ding2024quarvla, cai2023task2morph, li2024roboflamingo, shentu2024lcb, yu2024uniaff, li2024robonurse}, based on a pre-trained multi-modal large model capable of processing visual question answering (VQA), are fine-tuned on large-scale robot datasets. 
However, these models usually follow the structure of existing large language models, which fails to account for the unique characteristics of VLA tasks.

Additionally, the training of these models is conducted through imitation learning on expert data where execution is successful, and they are unable to learn from failure data collected automatically by robots. Although some approaches~\cite{song2024germ, chebotar2023qtransformer} have attempted to use reinforcement learning paradigms on datasets that include both successful and failed cases, they are still limited to smaller-scale models and have not applied these methods to multi-modal large language models.

\begin{figure*}[t]
    \centering
    \includegraphics[width=0.95\textwidth]{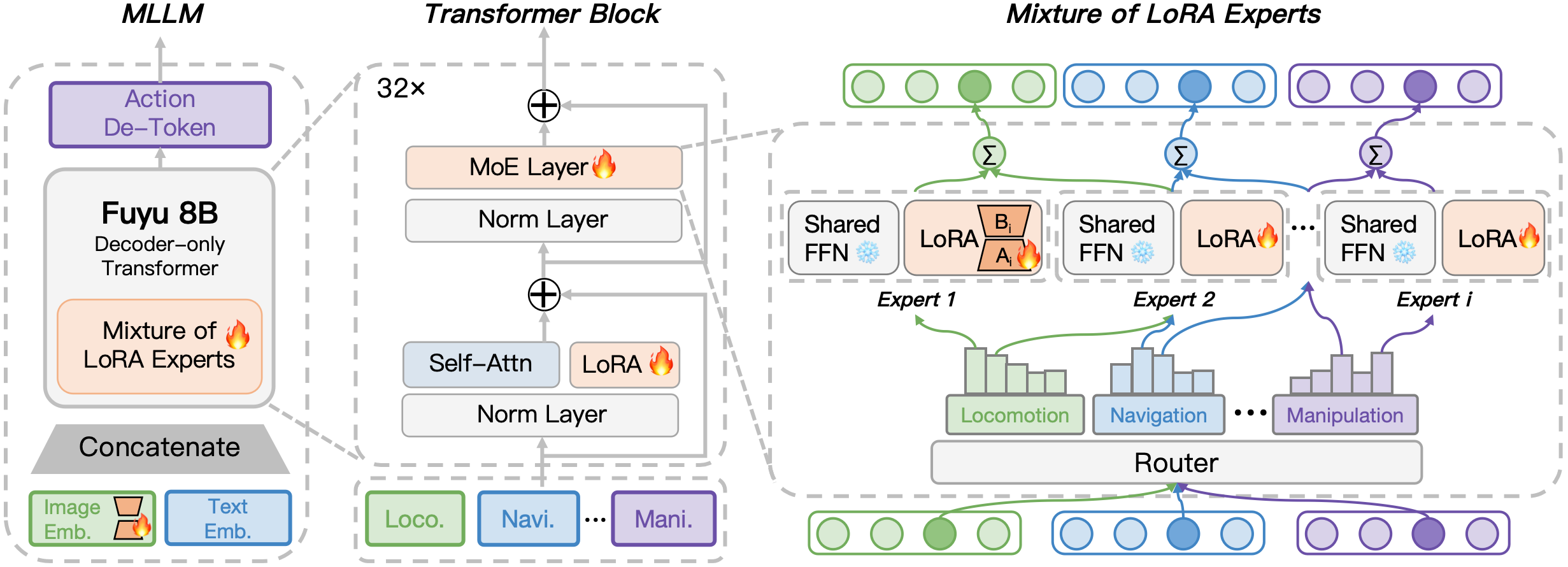}
    \caption{\textbf{The network architecture of \method.} This figure illustrates the architecture of \method, which uses a decoder-only transformer (Fuyu 8B~\cite{fuyu-8b}) integrated with a Mixture of LoRA Experts. 
    Tokens from different tasks such as locomotion, navigation, and manipulation are routed through a shared feed-forward network (FFN) with each expert dynamically selected by the router to provide the most relevant token-specific adaptation. The mixture-of-experts approaches allows for flexible token adaption within a single model.}
    \label{fig:architecture}
\end{figure*}

\subsection{Mixture-of-Experts in Multi-modal Large Language Models}
In the field of multi-modal large language models (MLLMs), there has been significant work attempting to incorporate the mixture-of-experts (MoE) architecture into the backbone of pre-trained language models to better accommodate tasks across multiple domains. These models can adaptive select different experts for processing based on the input data through a routing module, effectively preventing task conflicts that can arise during mixed-task training. 

MoE-LLaVA~\cite{lin2024moellava} duplicates the Feed Forward Network (FFN) from a dense MLLM multiple times to serve as the initialization for the MoE layers and only fine-tunes the MoE portion. Despite having fewer activated parameters than many larger-scale models, it achieves better performance on multiple VQA tasks. LLaVA-MoLE~\cite{chen2024llavamole} effectively combines the advantages of parameter-efficient fine-tuning and the MoE architecture by incorporating multiple low-rank matrices into the FFN. These models have achieved success in various domain-specific VQA tasks; however, there has been limited research delving into the advantages of the MoE architecture in robotic tasks.

\section{METHOD}

In this section, we first provide an introduction to some preliminaries, then elaborate on the structure of the proposed model, and finally present the training objective based on reinforcement learning (RL) that we utilize.
\subsection{Preliminaries} 
\label{sec:prelimanries}
\textbf{RL.}
The tasks that the robot executes can be viewed as a Markov decision process (MDP). At each time step, the robot is in a state $s$ and can take an action $a$. The transition dynamics is determined by the state transition function $T(s'|s,a)$, which is related to the environment, and a reward function $R(s,a)$ is obtained. The goal of RL is to learn a policy $\pi$ that selects actions $a$ to maximize the cumulative reward, where the dimensionality of a is~$\numDims$.
Value-based methods estimate the Q-function $Q(s,a)$ to evaluate the future discounted return of state-action pairs, expressed as $\sum_t \gamma^t R(s_t, a_t)$, and adopt $\pi (a|s) = {\operatorname{argmax}_{a}} Q(s,a)$ as the policy. The Q-function is updated through the iterative application of the Bellman operator
\begin{align}
\B^* Q(s_t,a_t) = R(s_t,a_t) + \gamma \max_{a_{t+1}}Q(s_{t+1}, a_{t+1}),
\vspace{-3pt}
\label{eq:bellman}
\end{align}
which is approximated via function approximation. 

\textbf{Mixture-of-experts (MoE).}
Unlike dense networks, the MoE architecture processes input by activating one or several submodules in a sparse layer. The selection of the activation modules is typically based on a learnable router layer. Considering an input $x$ for a network, the output $y$ of an MoE layer can be represented as
\begin{equation}
    y = \sum_{k=1}^{N}G(x)_k E_k(x),
\end{equation}
where $N$ is the total number of experts in the MoE layer, $E_k(x)$ is the output of the $k$-th expert. $G(x)_k$ represents the $k$-th element of the routing logits $G(x)$, which are typically obtained by applying a linear transformation to the input and then selecting the Top-$K$ elements with the highest values:
\begin{equation}
    G(x) = {\rm Softmax (Top}K(W_g x))
\end{equation}
In the MoE layer, the degree of sparsity of the network depends on the choice of $K$. When $K$ equals $N$, it signifies that all parameters in the layer are activated, whereas when $K$ is set to 1, only a single expert in the entire layer is activated to process the input. In large language models, MoE layers are often applied to the feed-forward network (FFN) portion of the model. Through this sparsity, the model can reduce computation while maintaining capabilities comparable to those of dense models of the same scale.
\subsection{Model Architecture}
\textbf{Overall structure.}
In Figure~\ref{fig:overview}, \method~enables training high-capacity transformer architectures on mixed quality data, consisting of narrow expert data and broad sub-optimal data. The image $I_{\rm RGB}$~and instruction $T_{\rm Inst}$~are separately processed into image and text tokens. 
The input tokens are concatenated and processed by the Fuyu 8B~\cite{fuyu-8b}, a decoder-only transformer that supports inputs of arbitrary resolution and number of images, as the MLLM backbone to generate action tokens.
It contains a sequence of transformer blocks (32 layers), each containing an MoE layer, preceded by normalization layers and self-attention layers with LoRA. 
Specifically, a Mixture of LoRA Experts module is designed to fine-tune the model. 
The probability of outputting an action can be expressed as:
\begin{equation}
    P_{\rm LM}(a_t | s_t) = \prod_{i=0}^{d_\mathcal{A}} P_{\rm LM}(a_t^i | I_{\rm RGB}; T_{\rm Inst}; a_t^{1:i-1}),
\end{equation}
where $t$ is a given time-step, $a_t^{i}$ represents the $i$-th action token output by the language model, and $a_t^{1:i}$ denotes a sequence composed of action tokens from the first dimension to the $i$-th dimension. 
Eventually, $a_t$ is de-tokenized into 12-dimensional discretized robot commands:
\begin{align}
\left[v_x, v_y, \omega_z, \theta_1, \theta_2, \theta_3, f, h_z, \phi, s_y, h_z^f, T \right]
\label{eq:dof}
\end{align}
Here, $v_x$, $v_y$, and $\omega_z$ represent the velocities along the x-axis, y-axis, and z-axis respectively. $\theta_1$, $\theta_2$, and $\theta_3$ indicate the gait pattern, $f$ denotes the frequency, $h_z$ represents the height of the robot, $\phi$ denotes the pitch angle, $s_y$ corresponds to the foot width, $h_z^f$ represents the foot height, and $T$ indicates the termination signal of the action.

\textbf{Mixture of LoRA experts.}
For the decoder layers in the language model, we employed the MixLoRA~\cite{li2024mixlora} approach for parameter-efficient fine-tuning and adjusted the FFN layers into MoE layers during the fine-tuning process. As shown in Figure~\ref{fig:architecture}, specifically during the fine-tuning process, each expert in the MoE layer contains an FFN with parameters shared among experts, and LoRA adapters are applied to each linear layer of the FFN. Therefore, each expert can be represented as:
\begin{equation}
    E_k(x) = (W_{down} + W_{down}^{{\rm LoRA}_k})f((W_{up} + W_{up}^{{\rm LoRA}_k})x)
\end{equation}
where $f(\cdot)$ is the activation function and $W_{layer}^{{\rm LoRA}_k}$ is the $k$-th LoRA adapter of a particular layer.

We also incorporate a single LoRA adapter into the attention module of each decoder layer. The shared FFN is initialized from the original parameters of the MLLM and the whole backbone model remains frozen. Only the LoRA adapters are fine-tuned during training. 

\subsection{Training Objective}
\label{objective}
\begin{figure}[t!]
    \centering
    \includegraphics[width=0.49\textwidth]{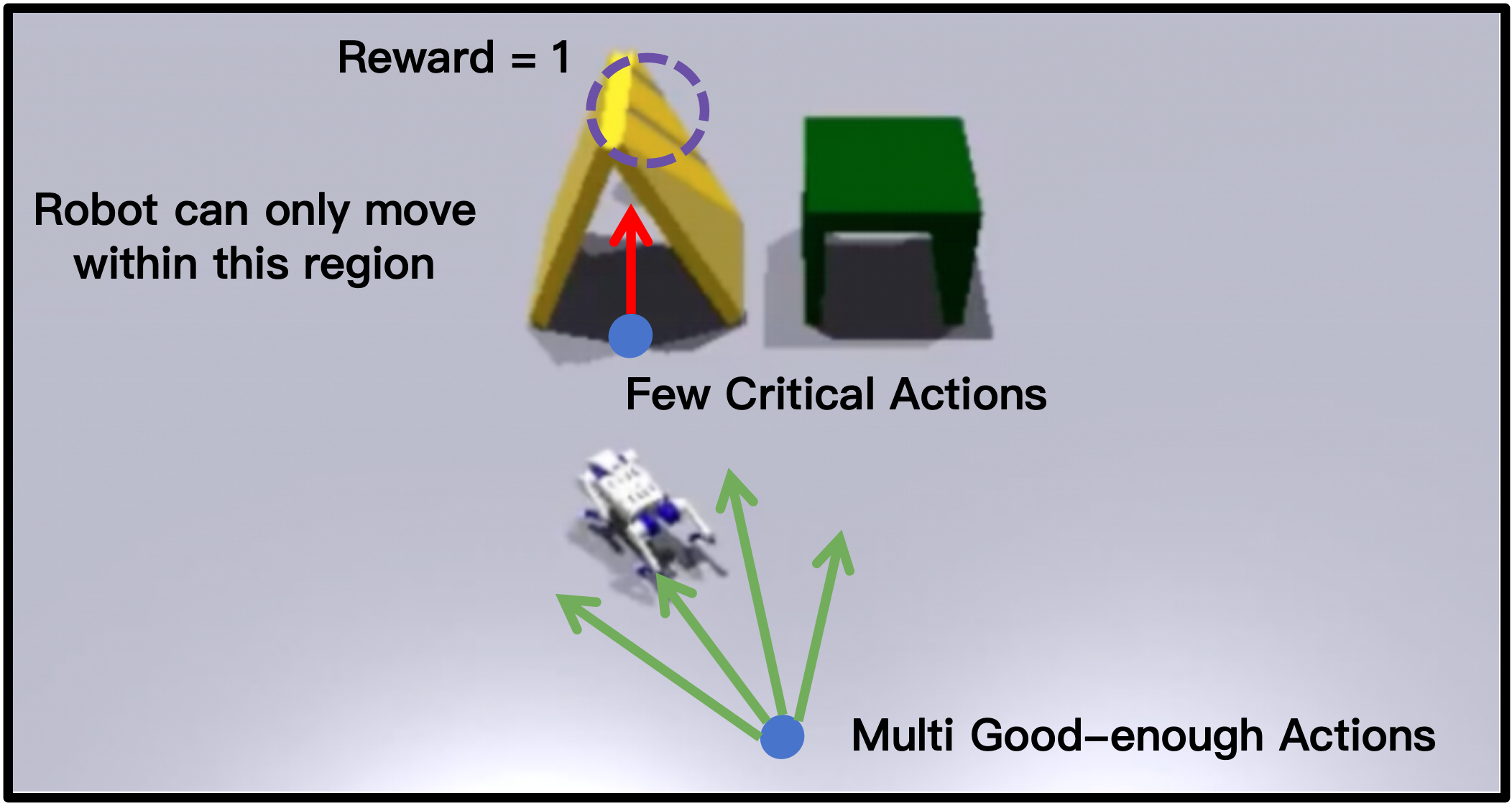}
    \caption{\textbf{The analysis of the structure of our task.} This figure illustrates the whole structure and intuition behind the critical points.}
    \label{fig:critical}
\end{figure}

As illustrated in Figure~\ref{fig:critical}, we identify 4 specific structural properties of the MDP in our task: 
(1) Horizon-independent returns. This property arises from our reward structure, where we assign a value of 1 solely upon successful task completion, with all other states yielding a reward of 0. 
(2) Limited critical points. The return of any trajectory can mostly be explained by the actions taken at a small fraction of critical states (\textit{e.g.}, crouching to avoid obstacles, dumping a load into a container). The lasting large proportion of states in a trajectory is not costly to recover after deviating from the optimal trajectory, this could be simply because there exists a large volume of near-optimal trajectories, or because at all but a few states, the volume of ``good-enough" actions is large and can be identified using reward information. 
(3) Long-horizon data. The trajectories in our task are generally long.
(4) Distribution shifts between the collected dataset and the evaluation phase.
These characteristics allow the policy to explore more freely and make sub-optimal decisions in non-critical states without compromising overall performance. 
In this scenario, offline RL algorithms can rapidly converge to good-enough actions in non-critical states, a capability that IL methods lack due to their inherent limitation of merely imitating the data distribution~\cite{whenshould}.

In this paper, we leverage the training loss function proposed in~\cite{song2024germ} and~\cite{chebotar2023qtransformer} to fully exploit the model’s capacity as a Q-function under the MDP of our task, and adapt it to better suit the fine-tuning process for large models. 
Under the setup of auto-regressive discrete Q-learning, where the policy treats each dimension of the action as a token discretized from continuous values and outputs them auto-regressively, the Bellman operator defined in~\eqref{eq:bellman} can be obtained with the following form:
\begin{align}
    Q(s_t, a_t^{1:i-1}, a_t^i) &= \notag\\
    &\begin{aligned}[t]
    \hspace{-8em}\begin{cases} 
     \underset{a_t^{i+1}}{\max}~Q(s_t,  a_t^{1:i}, a_t^{i+1}) & \text{if } i < d_\mathcal{A}\\
     R(s_t, a_t) + \gamma~\underset{a_{t+1}^1}{\max}~Q(s_{t+1}, a_{t+1}^1) & \text{if } i = d_\mathcal{A}
    \end{cases}
    \end{aligned}\label{eq:q_update}
\end{align}
\begin{table*}[t!]
\caption{\textbf{Overall performance.} We evaluate 25 episodes per task for the final checkpoint on 6 challenging tasks from QUARD and report the success rates. The tasks are divided into ``Easy", ``Medium" and ``Hard" according to their difficulty.}
\label{tab:exp_baseline}
\renewcommand\arraystretch{1.2}
\scriptsize
\centering
\begin{tabular}{cc|c|c|c|c|c|c|c|c }
\hline
\cline{4-10}
\multicolumn{3}{c|}{\thead{}} 
&\multicolumn{1}{c|}{\thead{Easy}} 
&\multicolumn{1}{c|}{\thead{Medium}}
& \multicolumn{4}{c|}{\thead{Hard}}
& \multicolumn{1}{c}{\thead{Average}}
\\
\multicolumn{2}{c|}{\thead{}} 
&\multicolumn{1}{c|}{\thead{Total\\Params}} 
&\multicolumn{1}{c|}{\thead{Distinguish}} 
&\multicolumn{1}{c|}{\thead{Go to}}
& \multicolumn{1}{c}{\thead{Go avoid}}
&\multicolumn{1}{c}{\thead{Go through} }
&\multicolumn{1}{c}{\thead{Crawl}} 
&\multicolumn{1}{c|}{\thead{Unload}}
& \multicolumn{1}{c}{\thead{}}
\\
\hline
{{CLIP}\cite{radford2021learning}}&
&86M &0.44 & 0.43 & 0.45& 0.19& 0 & 0 & 0.25\\

{VC-1}\cite{vc2023}&
&307M &0.46 &0.43 &0.45 &0.31 &0  &0 & 0.28\\
{QUART}\cite{ding2024quarvla}&
&8B &0.66  &0.60& 0.53& 0.41& 0.32& 0.12& 0.44\\

\textbf{\method~(ours)}&
&\textbf{9.82B} & \textbf{0.82} & \textbf{0.80} & \textbf{0.59} & \textbf{0.57} & \textbf{0.49} & \textbf{0.33}& \textbf{0.60}\\

\hline

\end{tabular}
\label{tb:result}
\end{table*}

\begin{table*}[t!]
\caption{\textbf{Ablation study}. We ablate 3 important designs in \method.}
\label{tab:exp_baseline}
\renewcommand\arraystretch{1.2}
\scriptsize
\centering
\begin{tabular}{cc|c|c|c|c|c|c|c|c }
\hline
\cline{4-9}
\multicolumn{3}{c|}{\thead{}} 
&\multicolumn{1}{c|}{\thead{Easy}} 
&\multicolumn{1}{c|}{\thead{Medium}}
& \multicolumn{4}{c|}{\thead{Hard}}
&\multicolumn{1}{c}{\thead{Average}}
\\
\multicolumn{2}{c|}{\thead{}} 
&\multicolumn{1}{c|}{\thead{S-Data}}  
&\multicolumn{1}{c|}{\thead{Distinguish}} 
&\multicolumn{1}{c|}{\thead{Go to}}
& \multicolumn{1}{c}{\thead{Go avoid}}
&\multicolumn{1}{c}{\thead{Go through} }
&\multicolumn{1}{c}{\thead{Crawl}} 
&\multicolumn{1}{c|}{\thead{Unload}}
&\multicolumn{1}{c}{\thead{}}
\\
\hline
{QUART}\cite{ding2024quarvla} & & N & 0.66 & 0.60 & 0.53 & 0.41 & 0.32 & 0.12 &0.44 \\ {w/o RL} & & N & 0.73 & 0.67 & 0.58 & 0.47 & 0.34 & 0.24 & 0.51 \\ {w/o MoE} & & Y & 0.70 & 0.63 & 0.55 & 0.45 & 0.37 & 0.18 &0.48 \\ {w/o S-Data} & & N & 0.78 & 0.69 & \textbf{0.61} & 0.53 & 0.45 & 0.28 &0.56 \\ \textbf{\method~(ours)} & & Y & \textbf{0.82} & \textbf{0.80} & 0.59 & \textbf{0.57} & \textbf{0.49} & \textbf{0.33} & \textbf{0.60}\\

\hline

\end{tabular}
\label{tb:ablation}
\end{table*}

We train on an offline dataset pre-collected by other behavior policies. To ensure training stability, it is necessary to incorporate some conservative terms in the training objective to constrain the policy to train on state-action pairs within the dataset scope, preventing excessive overestimation of Q-values outside the training set. Additionally, we aim to minimize the probability of invalid action tokens, making the RL loss function defined as:
\begin{align}
    \hspace{-2pt} \mathcal{L}_{RL} &= ~\frac{1}{2}~ \mathbb{E}_{s \sim \mathcal{D},a \sim \pi_\beta(a|s)} \left[\left(Q(s,a) - \mathcal{B}^* Q^{k}(s,a)\right)^2\right] \notag \\
    &\quad + \alpha \cdot \frac{1}{2}\mathbb{E}_{s \sim \mathcal{D},a \sim \tilde{\pi}_\beta(a|s)} \left[(Q(s,a) - 0) ^2\right], \label{eq:opt_obj}
\end{align}
where $Q(s, a) = \sigma (P_{\rm LM}(a | s))$ and $\sigma$ is the sigmoid activation, which ensures that the values of the Q-function fall within the range $(0, 1)$. $\pi_\beta(a|s)$ is the behavior policy distribution of the action during data collection, $\alpha$ is a factor that modulates the strength of the conservative regularization. $\tilde{\pi}_\beta(a|s) = \frac{1}{Z(s)} (1.0 - \pi_\beta(a|s))$ is the distribution over all actions which have a very low density under $\pi_\beta(a|s)$, where ${Z(s)}$ is a normalizing term within the distribution. Since the output $P_{\rm LM}$ of the large model denotes the distribution over the entire vocabulary of the language model, it also includes tokens that do not represent a valid action.

Additionally, in order to prevent certain experts in the MoE layer from being frequently selected, leading to an imbalance in workload, we adopt an auxiliary loss used in~\cite{fedus2022switchtransformers} to encourage the workload balance across experts:
\begin{equation}
    \mathcal{L}_{MoE} = \frac{1}{N}\sum_{k=1}^N f_k P_k,
\end{equation}
where $f_k = \frac{1}{T} \sum_{x \in \mathcal{D}} 1 \{\arg\max p(x) = k\}$ denotes the fraction of tokens dispatched to the $k$-th expert, and $P_k = \frac{1}{T} \sum_{x \in \mathcal{D}} p_k(x)$ is the fraction of the router probability allocated for the $k$-th expert.

The final loss function in the training process is set as:
\begin{equation}
    \mathcal{L} = \mathcal{L}_{RL} + \beta \mathcal{L}_{MoE},
\end{equation}
and we choose the coefficient as $\alpha=0.5$ and $\beta=0.002$.

\section{EXPERIMENTS}
We concentrated experiments on evaluating the multi-task ability, generalization capabilities, and real-world deployment and aim to answer the following questions: 
\textbf{Q1.} Why is \method~preferable than existing VLA methods?
\textbf{Q2.} Why were the design choices? 
\textbf{Q3.} Does \method~learn a superior policy that handles real robot tasks in 6 different scenes with limited real-world data?

\subsection{Experiments Setup}
\textbf{Simulation environment.}
We use NVIDIA Isaac Gym~\cite{makoviychuk2021isaac} to collect the trajectories of the robot executing different tasks in parallel.

\textbf{Real-world setting.}
The robot used to collect the trajectory data is Unitree Go2, which is a quadruped robot with 12 joints. The command output is sent to the low-level command tracking controller, which will be discussed below, to generate the actual joint action of the robot. 
The perception data is provided by a RealSense d435 camera installed in the front of the robot. The real-world experiment was executed by manually manipulating the quadruped robot. 

\textbf{Mixed-quality Datasets.}
The offline dataset used in our experiment includes two categories: demonstrations and sub-optimal data.
Demonstrations correspond to successful tasks, which consist of 6 types of tasks (``Distinguish'', ``Go to'', ``Go through'', ``Go avoid'', ``Unload'', ``Crawl'') with a total of 1,822,405 vision-language-action sets, all sourced from human demonstration data in QUARD \cite{ding2024quarvla}.
This part of sub-optimal data comprises 4 types of tasks (``Distinguish'', ``Go to'', ``Go through'', ``Go avoid'') and includes a total of 440,732 vision-language-action sets with all data sourced from QUARD-Auto \cite{song2024germ}.

\textbf{Low-level controller.}
We use the RL policy trained in walk-these-ways~\cite{walktheseway} as the low-level controller for our robot. 
In walk-these-ways, a controller capable of controlling various gaits and body instructions was trained using RL methods with a mix of different reward functions related to diverse gait and body state commands. 
The policy was trained in parallel across a large number of simulation environments and different environmental parameters. 
To enhance the ability to execute a wide range of commands, a variety of commands are randomized within a specific range, allowing for comprehensive training across different operations.
By adjusting these command parameters, complex behaviors can be generalized during inference.

\textbf{Baseline.}
We compared our method with three previous state-of-the-art approaches on the quadruped robot VLA task: QUART~\cite{ding2024quarvla}, VC-1~\cite{vc2023}, and CLIP~\cite{radford2021learning}. 
To ensure a comprehensive and thorough evaluation, we maintained the same implementation as in~\cite{ding2024quarvla}, which enabled these methods to work as a VLA model. 

\begin{figure}[t!]
    \centering
    \includegraphics[width=0.49\textwidth]{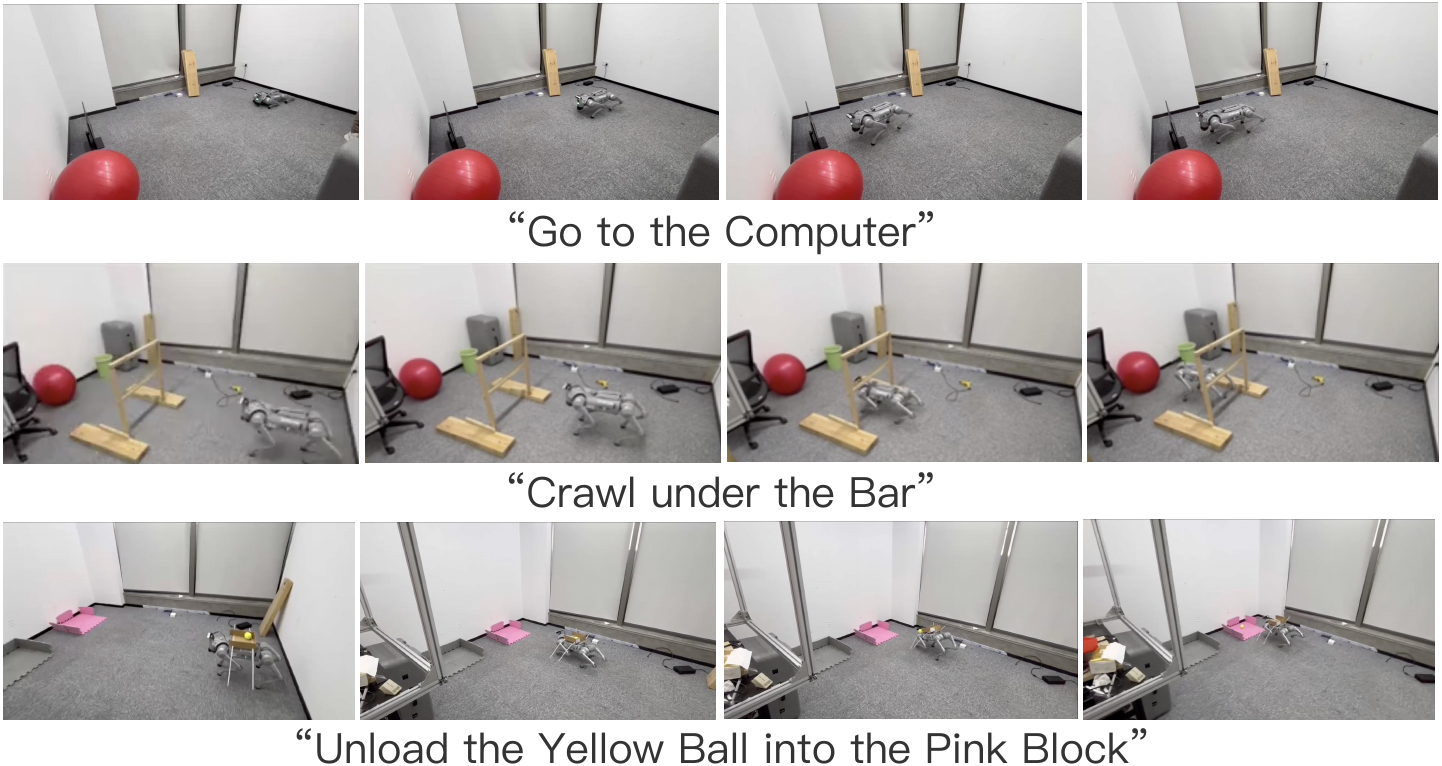}
    \caption{\textbf{Real-world experiments}. These images showcase the robot successfully performing various tasks, involving navigation (\textbf{``Go to"}), adjusting body posture during locomotion (\textbf{``Crawl"}), and whole body manipulation (\textbf{``Unload"}).}
    \label{fig:real}
\end{figure}

\textbf{Training details.}
For the training on expert data, our model was trained using 8 NVIDIA A100 GPUs over 3 epochs, which required approximately 100 hours. The batch size per device was set to 16. 
For the mixed-quality training, we maintained the same hyper parameter settings, with the total training time amounting to approximately 125 hours.

To ensure a fair comparison between IL and RL methods, we exclusively used failure data from the QUARD-Auto dataset to construct the mixed-quality training set, without incorporating any successful data.

\textbf{Evaluation Details.}
We conducted a comprehensive and robust series of experiments. 
We define a comprehensive set of evaluation tasks that covers various axes of generalization, such as visual (unseen objects, distractor objects), motion (positions of the target objects and the robot), locomotion (different gaits/ speed), and semantic (unseen instructions) generalization.
To rigorously test the overall performance and generalization capabilities of~\method, these out-of-distribution scenes and instructions were systematically generated together with in-distribution scenes.

\subsection{Main Results in Simulation}
Table~\ref{tb:result} shows that \method~consistently outperforms all baseline methods across the majority of tasks, achieving improvements of up to 20\%.
The exceptional success rate of \method~in the ``distinguish" and ``go to" tasks highlights the strong foundational perception capabilities of MLLM. Its progress in tasks like ``go through" and ``go avoid" further reflects an understanding of spatial relationships. 
Additionally, the success in more complex tasks such as ``crawl" and ``unload" demonstrates the ability of whole-body manipulation, which is enabled by the implicit learning of interdependencies between different action dimensions through the use of a transformer.
The consistently high success rate across all tasks is attributed to the mixture of robot experts, where each expert specializes in particular action dimensions and skills, allowing adaptation to various tasks.
Moreover, training on tasks with sub-optimal data reveals potential for even greater success rate gains. 
Learning the ability to identify good-enough choice from the reward information is easy, and luckily our task setting allows a large volume of good-enough action, as we discussed in section~\ref{objective}.
Given that the evaluation includes out-of-distribution scenes, such robust performance further confirms the generalization capabilities of~\method, which can be attributed to the common sense of pretrained MLLM.

\subsection{Ablation Study}
Table~\ref{tb:ablation} presents a detailed summary of the ablation studies performed on three key components of \method. 
The findings are as follows:
1. The ablation study of the RL loss demonstrates a significant performance boost with its inclusion. Notably, even in scenarios with only expert data, the success rate of the offline RL algorithm surpasses that of IL methods, which also benefit from RL choice as mentioned before.
2. The ablation study of the MoE modules shows that the MoE modules balance computational cost and performance by activating only a portion of the parameters during inference. While the number of activated parameters remains comparable to that of a standard model, the multiple experts adapt to different task tokens, thereby improving the success rate in multi-task learning.
3. The ablation study of sub-optimal data shows that the inclusion of sub-optimal data improves the performance of~\method~through RL. 
However, the overly sparse reward function introduces learning challenges, which leads to performance degradation in ``Go avoid" task.

\subsection{Real-world Results}
Furthermore, we tested the performance of MoRE in real-world scenarios.
We further fine-tuned our model on a small real-world dataset and obtained an instance that can be deployed in real-world settings. We conducted experiments in several scenarios and achieved promising results. As shown in Figure~\ref{fig:real}, \method~successfully completed tasks such as ``go to the computer", ``crawl under the bar", and ``unload the yellow ball" in real-world scenarios, despite our real-world dataset only containing tasks like ``go to the colored ball” with several different colors or ``distinguish the letter" with different letters on the box. This demonstrates the strong generalization capability of our model to real-world scenarios after training with simulation data.

\section{CONCLUSIONS}
In this paper, we present~\method, a large-scale VLA model to unlock scalability in reinforcement learning.
By integrating a mixture of robot experts into the VLA model, \method~demonstrates strong effectiveness in multi-task learning. Furthermore, utilizing the VLA model as a Q-function enables \method~to adopt an optimal training pipeline. Lastly, through the incorporation of mixed-quality data, \method~gains the ability to learn from sub-optimal data, further enhancing its performance.
\method~achieves high success rate across tasks of different levels of difficulties and further behaves generalization skills on out-of-distribution scenes.
The results are validated in both simulation and real-world robot settings.

\section*{ACKNOWLEDGMENTS}
This work was supported by the National Science and Technology Innovation 2030 - Major Project (Grant No. 2022ZD0208800), and NSFC General Program (Grant No. 62176215).






%


\newpage
\bibliographystyle{IEEEtranS}
\bibliography{root}



\end{document}